\documentclass{article}
\pdfoutput=1

\usepackage{microtype}
\usepackage{graphicx}
\usepackage{booktabs} %

\usepackage{caption,subcaption}
\usepackage{amsmath,amssymb,color}
\usepackage{multirow}
\usepackage{enumitem}
\graphicspath{{figures/}} 
\usepackage[accepted]{icml2020}

\icmltitlerunning{Guided Learning of Nonconvex Models through Successive Functional Gradient Optimization}

\makeatletter
\newcommand\tightpara{\@startsection{paragraph}{4}{\z@}{0.2ex plus
   0ex minus 0ex}{-1em}{\normalsize\bf}}
\newcommand\normalpara{\@startsection{paragraph}{4}{\z@}{1.5ex plus
   0.5ex minus .2ex}{-1em}{\normalsize\bf}}    
\makeatother 
\newcommand{\mypara}{\tightpara}

\begin{document}

\twocolumn[
\icmltitle{Guided Learning of Nonconvex Models through Successive Functional Gradient Optimization}

\icmlsetsymbol{equal}{*}

\begin{icmlauthorlist}
\icmlauthor{Rie Johnson}{one}
\icmlauthor{Tong Zhang}{two}
\end{icmlauthorlist}

\icmlaffiliation{one}{RJ Research Consulting, Tarrytown, New York, USA}
\icmlaffiliation{two}{Hong Kong University of Science and Technology, Hong Kong}

\icmlcorrespondingauthor{Rie Johnson}{riejohnson@gmail.com}
\icmlcorrespondingauthor{Tong Zhang}{tongzhang@tongzhang-ml.org}

\icmlkeywords{Deep learning, Neural networks, Self-distillation, Training method, Regularization, 
              Image classification, Sentiment classification}

\vskip 0.3in
]

\printAffiliationsAndNotice{}  %

\begin{abstract}
This paper presents a framework of successive functional gradient optimization for training nonconvex models 
such as neural networks, where training is driven by mirror descent in a function space.  We provide 
a theoretical analysis and empirical study of the training method derived from this framework.  
It is shown that the method leads to better performance than that of standard training techniques.  
\end{abstract}

\newcommand{\Algm}{Alg-$m$}
\newcommand{\AlgL}{Alg$L$}
\newcommand{\Algsq}{Alg-sq}
\newcommand{\ourItes}{stages}
\newcommand{\ourIte}{stage}

\newcommand{\cmt}[1]{ !({\em #1 })! }

\newcommand{\netfunc}{ f }
\newcommand{\net}[2]{ \netfunc(#1;#2) }
\newcommand{\tarffx}[1]{ \netfunc_{#1}^* }
\newcommand{\hessian}[1]{{\rm H}\left(#1\right)}
\newcommand{\reg}{R}

\newcommand{\meanOverX}[2]{ \big \langle #2 \big \rangle_{(x,y)\in #1}}
\newcommand{\meanOverXsmall}[2]{ \langle #2 \rangle_{(x,y)\in #1}}
\newcommand{\meanOverXp}[2]{\meanOverX{#1}{#2}}
\newcommand{\trnset}{S}
\newcommand{\meanOverSp}[1]{ \meanOverX{\trnset}{#1}}
\newcommand{\meanOverS}[1]{\meanOverX{\trnset}{#1}}
\newcommand{\batch}{B}
\newcommand{\meanOverBp}[1]{ \meanOverX{\batch}{#1}}
\newcommand{\meanOverB}[1]{\meanOverX{\batch}{#1}}

\newcommand{\prm}{\theta}
\newcommand{\Bref}{h} %
\newcommand{\Bdivx}[1]{D_{#1}}
\newcommand{\Bdiv}{\Bdivx{\Bref}}   %
\newcommand{\loss}{L}
\newcommand{\lossy}{\loss_y}
\newcommand{\myloss}[1]{\loss\left( #1, y \right)}
\newcommand{\Bdivloss}{\Bdivx{\lossy}}

\newcommand{\assign}{ \leftarrow }
\newcommand{\tar}{q}

\newcommand{\ssa}{\alpha}

\newcommand{\numCls}{C}
\newcommand{\idxCls}{c}
\newcommand{\prob}{\sigma}

\newcommand{\baseInFig}{base}
\newcommand{\baseInTable}{base model}
\newcommand{\baseLoop}{base-loop}
\newcommand{\baseLam}{base-$\lam/\ssa$}
\newcommand{\iniRandom}{ini:random}
\newcommand{\iniBase}{ini:base}
\newcommand{\fcsmeta}{V} 

\newcommand{\Real}{{\bb R}}
\newcommand{\lam}{\lambda}
\newcommand{\rE}{{\mathbb E}}

\newtheorem{definition}{Definition}[section]
\newtheorem{lemma}{Lemma}[section]
\newtheorem{corollary}{Corollary}[section]
\newtheorem{theorem}{Theorem}[section]
\newtheorem{proposition}{Proposition}[section]
\newtheorem{assumption}{Assumption}[section]
\newcommand{\BlackBox}{\rule{1.5ex}{1.5ex}} 
\newenvironment{proof}{\par\noindent{\bf Proof\ }}{\hfill\BlackBox\\[2mm]}
\newcommand{\reglossx}[1]{\ell_{#1}}
\newcommand{\tarff}{ \netfunc^* }
\newcommand{\tT}{ T }
\newcommand{\ours}{GULF}
\newcommand{\oursL}{\ours2}   %
\newcommand{\ourssq}{\ours1}  %

\newcommand{\algm}{Algorithm \ref{alg:Algm}}
\newcommand{\algL}{Algorithm \ref{alg:AlgL}}
\newcommand{\algsq}{Algorithm \ref{alg:Algsq}}
\newcommand{\algLs}{Alg.\ref{alg:AlgL}} %
\newcommand{\algsqs}{Alg.\ref{alg:Algsq}} %
\newcommand{\algboth}{Algorithms \ref{alg:Algsq} and \ref{alg:AlgL}}

\newcommand{\netprmLong}{\net{\prm}{x}}
\newcommand{\netprm}{\netfunc_{\prm}} %
\newcommand{\netprmtLong}{\net{\prm_t}{x}}
\newcommand{\netprmt}{\netfunc_{\prm_t}} %
\newcommand{\ssai}{\gamma}

\newcommand{\lr}{\eta}
\newcommand{\prmx}[1]{\prm^{#1}}
\newcommand{\netprmx}[1]{\netfunc_{\prmx{#1}}}
\newcommand{\netprmj}{\netprmx{j}}
\newcommand{\netprmn}{\netprmx{n}}

\newcommand{\reglossa}{\reglossx{\ssa}(\prm)}

\section{Introduction}
\label{sec:intro}

This paper presents a new framework to train nonconvex models such as
neural networks. The goal is to learn a vector-valued function 
$\net{\prm}{x}$ that predicts an output $y$ from input $x$,
where $\prm$ is the model parameter. 
For example, 
for $K$-class classification where $y \in \{1,2,\ldots,K\}$, 
$\net{\prm}{x}$ is $K$-dimensional, and it can be linked to 
conditional probabilities via the soft-max logistic function.  
Given a set of training data $\trnset$, %
the standard method for solving this problem is to use 
stochastic gradient descent (SGD) for finding a parameter that minimizes 
on $\trnset$ %
a loss function $\loss(\net{\prm}{x},y)$ with a regularization term $\reg(\prm)$: 
$
             \min_\prm \left[ 
                \frac{1}{|\trnset|}\sum_{(x,y)\in\trnset} \loss(\net{\prm}{x},y)
                + \reg(\prm)
             \right]. 
$

In this paper, we consider a new framework %
that %
{\em guides training through successive functional gradient descent} so that 
training proceeds with alternating the following: 
\begin{itemize}[itemsep=1pt,topsep=1pt]
\item Generate a guide function so that it is ahead (but not too far ahead) 
      of the current model with respect to the minimization of the loss.  
      This is done by functional gradient descent.   
\item `Push' the model towards the guide function.  
\end{itemize}
Our original motivation was functional gradient learning of additive models in 
{\em gradient boosting} \cite{frie:01}. 
In our framework, essentially, training proceeds with repeating a local search, 
which limits the searched parameter space to the functional neighborhood of the
current parameter at each iteration, instead of searching the entire space at once as the standard method does.   
This is analogous to $\varepsilon$-boosting 
where the use of a very small step-size (for successively expanding the ensemble of weak functions)
is known to achieve better generalization \cite{frie:01}.

For measuring the distances between models, %
we use the Bregman divergence (see e.g., \cite{convex15})
by applying it to the model output.  
Given a convex function $\Bref$, the Bregman divergence $\Bdiv$ is defined by 
\begin{equation}
  \Bdiv(u,v) = \Bref(u) - \Bref(v) - \nabla \Bref(v)^\top (u-v). \label{eq:bdiv}
\end{equation}
This is the difference between $\Bref(u)$ and the approximation of $\Bref(u)$ 
based on the first-order Taylor expansion around $v$.  This means that when $u-v$ is small, 
\begin{equation}
  \Bdiv(u,v) \approx \frac{1}{2} (u-v)^\top(\hessian{\Bref(v)})(u-v) , \label{eq:bdivApprox}
\end{equation}
where $\hessian{h(v)}$ denotes the Hessian matrix of $h$ with respect to $v$. 
Therefore, use of the Bregman divergence has the beneficial effect of utilizing the second-order information. 

We show that the parameter update rule of an induced method 
generalizes that of {\em distillation} \cite{distill14}. 
That is, our framework subsumes iterative {\em self-distillation} 
as a special case.  

Distillation was originally proposed to {\em transfer knowledge} 
from a high-performance but cumbersome model to a more manageable model. 
Various forms of self-distillation, 
which applies distillation to the models of the same architecture, 
has been empirically studied 
\cite{xl19,generation19,LZG18,Ban18,onlinedistill18,deepMutual18,meanTeachers17,gift17}.
One trend is to add to the original scheme, e.g., adding a term to the update rule, 
data distortion/division, more models for mutual learning, and so forth. 
However, we are not aware of any work on theoretical understanding such as
a convergence analysis of the basic self-learning scheme.  

Our theoretical analysis of the proposed framework provides a new functional gradient view of 
self-distillation, and we show a version of the generalized self-distillation
procedure converges to a stationary point of a regularized loss function. 
Our empirical study shows that the
iterative training of the derived method goes through a `smooth path'
in a restricted region
with good generalization performance.  
This is in contrast to standard training, where the entire (and therefore much larger) 
parameter space is directly searched, and thus complexity may not be well controlled.  
\mypara{Notation}
$\nabla h(v)$ denotes the gradient of a scalar function $h$ with respect to $v$. 
We omit the subscript of $\nabla$ when the gradient is with respect to 
the first argument, e.g., we write $\nabla \net{\prm}{x}$ for $\nabla_\prm \net{\prm}{x}$.  
$\hessian{h(v)}$ denotes the Hessian matrix of a scalar function $h$ with respect to $v$.  
We use $x$ and $y$ for input data and output data, respectively.  
We use $<>$ to indicate the mean, e.g.,
$
  \meanOverXsmall{\trnset}{F(x,y)}
  = \frac{1}{|\trnset|}\sum_{(x,y)\in\trnset} F(x,y). 
$
$\loss(u,y)$ is a loss function with $y$ being the true output. 
We also let $\loss_y(u) = \loss(u,y)$ when convenient.

\newcommand{\pp}{p}
\newcommand{\ff}{f}
\newcommand{\ffstar}{f^*}

\section{Guided Learning through Successive Functional Gradient Optimization} 
\label{sec:theory}

In this section, after presenting the framework in general terms, 
we develop concrete algorithms and analyze them.  

\subsection{Framework} 

We first describe the framework in general terms so that the models to be 
trained are not limited to parameterized ones. 
Let $\ff$ be the model we are training.  
Starting from some initial $\ff$, training proceeds by repeating the following: 
\begin{enumerate}[itemsep=1pt,topsep=1pt]
  \item Generate a {\em guide function} $\ffstar$ by applying 
        functional gradient descent for reducing the loss to the current model $\ff$, 
        so that $\ffstar$ is an improvement over $\ff$ in terms of loss 
        but not too far from $\ff$.  
  \item Move the model $\ff$ in the direction of the guide function $\ffstar$
        according to some distance measure. 
\end{enumerate}

We use the Bregman divergence $\Bdiv$, defined in \eqref{eq:bdiv}, 
for representing the distances between models.
\mypara{Step 1: Guide going ahead}
We formulate Step 1 %
as 
\begingroup
\thinmuskip=0mu
\medmuskip=0mu
\thickmuskip=0mu
\begin{align}
    \ffstar(x,y) &:= \arg\min_\tar \left[ 
        \Bdiv(\tar,\ff(x)) + \ssa \nabla \lossy(\ff(x))^\top \tar
    \right ] , \label{eq:mirror0}  
\end{align}
\endgroup
where $\ssa$ is a meta-parameter.  
The second term pushes the guide function towards the direction of reducing loss, 
and the first term pulls back the guide function towards the current model $\ff$.  
Thus, $\ffstar$ is ahead of $\ff$ but not too far ahead.  
Note that we use the knowledge of the true output $y$ here; therefore, $\ffstar$ takes 
$y$ as the second argument.  
The function value for each data point $(x,y)$ can be found approximately 
by solving the optimization problem by SGD if there is no analytical solution. 
Also, this formulation is equivalent to finding $\ffstar$ such that 
\begin{align}
  \nabla \Bref(\ffstar(x,y)) = \nabla \Bref(\ff(x)) - \ssa \nabla \lossy(\ff(x)) . 
  \label{eq:mirror1} 
\end{align}
This is mirror descent (see e.g., \cite{convex15}) performed in a function space.  

Due to the relation of the Bregman divergence to the Hessian matrix stated in \eqref{eq:bdivApprox}, 
\eqref{eq:mirror0} implies that 
\begin{align}
  \ffstar(x,y) \approx \ff(x) - \ssa (\hessian{\Bref(\ff(x))})^{-1} \nabla \lossy(\ff(x)) .
  \label{eq:step}
\end{align}

Therefore, if we set $\Bref(f)=\lossy(f)$, \eqref{eq:step} becomes 
\begin{align}
  \ffstar(x,y) \approx \ff(x) - \ssa (\hessian{\lossy(\ff(x))})^{-1} \nabla \lossy(\ff(x)), 
  \label{eq:newton}
\end{align}
which is approximately a second-order functional gradient step 
(one step of the relaxed Newton method) with step-size $\ssa$ for minimizing the loss.  

If we set $\Bref(f)=\frac{1}{2}\|f\|^2$, then the optimization problem \eqref{eq:mirror0} 
has an analytical solution
\[
  \ffstar(x,y) = \ff(x) - \ssa \nabla \lossy(\ff(x)), 
\]
which is a first-order functional gradient step with step-size $\ssa$
for minimizing the loss.  

\mypara{Taking $m$ steps in Step 1}
For further generality, let us also consider $m$ steps of functional gradient descent by 
extending $\ffstar$ in \eqref{eq:mirror0} to $\ffstar_m$ recursively defined as follows. 
\begin{align*}
    \ffstar_0(x,y) &:= \ff(x) \\
    \ffstar_{i+1}(x,y) &:= \arg\min_\tar \left[ 
        \Bdiv(\tar,\ffstar_i(x)) + \ssa \nabla \lossy(\ffstar_i(x))^\top \tar
    \right ]. %
\end{align*}
Then, 
in parallel to \eqref{eq:step}, we have 
\begin{align*}
  \ffstar_{i+1}(x,y) \approx \ffstar_i(x) - \ssa (\hessian{\Bref(\ffstar_i(x))})^{-1} \nabla \lossy(\ffstar_i(x)) .
\end{align*}

\mypara{Step 2: Following the guide}  
Using the Bregman divergence $\Bdiv$, we formulate Step 2 above as an update of the model $\ff$ 
to reduce 
\begin{align}
    \meanOverS{ \Bdiv(\ff(x), \ffstar(x,y)) } + \reg(\ff)
    \label{eq:fit}
\end{align}
so that the model $\ff$ approaches the guide function $\ffstar$ 
in terms of the Bregman divergence.  
$\reg(\ff)$ is a regularization term.  

\mypara{Parameterization}
Although there can be many variations of this scheme, in this work, 
we parameterize the model $\ff$ so that we can train neural networks.  
Thus, we replace $\ff(x)$ by $\net{\prm}{x}$ with parameter $\prm$.  
This does not affect Step 1, and to reduce \eqref{eq:fit} in Step 2, 
we repeatedly update the model parameter $\prm$ by descending the stochastic gradient 
\begin{align}
\nabla_\prm \left[ 
   \meanOverB { \Bdiv(\net{\prm}{x}, \ffstar(x,y)) } + \reg(\prm)
\right] , 
\end{align}
where $\batch$ is a mini-batch sampled from a training set $\trnset$.  

\subsection{Algorithms}

Putting everything together, we obtain \algm, 
which performs mirror descent in a function space 
in Line 3.  
We call it (and its derivatives) a method of 
{\em GUided Learning through successive Functional gradient optimization (\ours)}.   
We now instantiate function $\Bref$ used by the Bregman divergence $\Bdiv$ 
to derive concrete algorithms.  In general we allow $\Bref$ to vary
for each data point. That is, it may depend on $(x,y)$. 
Here we use two functions discussed above, which correspond to 
the first-order and the second-order methods, respectively; 
however, note that choice of $\Bref$ is not limited to these two.  

\begin{algorithm}[h]
\begin{small}
   \caption{
      \ours\ in the most general form. 
      {\bfseries Input:} %
       $\prm_0$, 
       training set $\trnset$. %
       Meta-parameters: $m$, $\ssa$, $\tT$. 
      {\bfseries Output:} $\prm_\tT$. 
   }
   \label{alg:Algm}
\begin{algorithmic}[1]
   \STATE $\prm \assign \prm_0$
   \FOR{$t=0$ {\bfseries to} $\tT-1$}  %
      \STATE Define             
               $\tarffx{m}$ by: $\tarffx{0}(x,y) := \net{\prm_t}{x}$,
               $\tarffx{i+1}(x,y) := $\\ $~~~~~~~~~~~~\arg\min_\tar \left[ 
                  \Bdiv(\tar,\tarffx{i}(x,y)) + \ssa \nabla \lossy(\tarffx{i}(x,y))^\top \tar
                \right ]$        
      \REPEAT 
         \STATE Sample a mini-batch $\batch$ from $\trnset$.
         \STATE  Update $\prm$ by descending the stochastic gradient \\
             $\nabla_\prm \left[ 
               \meanOverB{ \Bdiv(\net{\prm}{x}, \tarffx{m}(x,y)) } 
               + \reg(\prm)
             \right]$ \\
             $~~~~$for optimizing \\
             $Q_t(\prm) :=     \meanOverS{ \Bdiv(\net{\prm}{x}, \tarffx{m}(x,y)) } 
               + \reg(\prm)$.
      \UNTIL{some criteria are met}
      \STATE $\prm_{t+1} \assign \prm$
   \ENDFOR
\end{algorithmic}
\end{small}
\end{algorithm}
\mypara{\ourssq\ (1st-order, \algsq)} 
With $\Bref(u)=\frac{1}{2}\| u \|^2$, %
we obtain Algorithm \ref{alg:Algsq}.  Derivation is straightforward.  
This algorithm performs $m$ steps of first-order functional gradient descent (Line 3)
to push the guide function ahead of the current model 
and then let the model follow the guide by reducing the 2-norm between them.   
\begin{algorithm}[h]
\begin{small} 
   \caption{\ourssq\ ($\Bref(u)=\frac{1}{2}\|u\|^2$): 
       {\bfseries Input:} %
       $\prm_0$, 
       training set $\trnset$. %
       Meta-parameters: $m$, $\ssa$, $\tT$.    
       {\bfseries Output:} $\prm_\tT$. 
   }
   \label{alg:Algsq}
\begin{algorithmic}[1]
   \STATE $\prm \assign \prm_0$
   \FOR{$t=0$ {\bfseries to} $\tT-1$}   
      \STATE Define $\tarffx{m}$ by: $\tarffx{0}(x,y) = \net{\prm_t}{x}$, \\
      $~~~~~~~~~~~~~~~~~~~~~~~~$$\tarffx{i+1}(x,y) = \tarffx{i}(x,y) - \ssa \nabla \lossy(\tarffx{i}(x,y))$   
      \REPEAT
         \STATE Sample a mini-batch $\batch$ from $\trnset$. 
         \STATE Update $\prm$ by descending the stochastic gradient \\
             $~~~~~~\nabla_\prm \left[ 
               \meanOverB{ \frac{1}{2}\| \net{\prm}{x} - \tarffx{m}(x,y) \|^2 } 
               + \reg(\prm)
             \right]$ 
      \UNTIL{some criteria are met}
      \STATE $\prm_{t+1} \assign \prm $
   \ENDFOR
\end{algorithmic}
\end{small}
\end{algorithm}
\begin{algorithm}[h]
\begin{small}
   \caption{\oursL\ ($\Bref(p)=\lossy(p)$): 
       {\bfseries Input:} %
       $\prm_0$, 
       training set $\trnset$. %
       Meta-parameters: $\ssa \in (0,1)$, $\tT$.   
       {\bfseries Output:} $\prm_\tT$. 
       Notation: $\netprm=\netprmLong$ and $\netprmt=\netprmtLong$.
   }
   \label{alg:AlgL}
\begin{algorithmic}
   \STATE $\prm \assign \prm_0$
   \FOR{$t=0$ {\bfseries to} $\tT-1$}  
      \REPEAT 
         \STATE Sample a mini-batch $\batch$ from $\trnset$. 
         \STATE Update $\prm$ by descending the stochastic gradient \\
             $\nabla_\prm \left[ 
               \meanOverB{ \Bdivx{\lossy}(\netprm, \netprmt) + \ssa \nabla \lossy(\netprmt)^\top \netprm }               
               + \reg(\prm)
             \right]$ 
      \UNTIL{some criteria are met}
      \STATE $\prm_{t+1} \assign \prm$
   \ENDFOR
\end{algorithmic}
\end{small}
\end{algorithm}
\mypara{\oursL\ (2nd order, \algL)} 

We consider the case of $\Bref(p)=\lossy(p)$ (i.e., $\Bref$ returns loss given prediction $p$).  
\eqref{eq:newton} has shown that in this case 
Step 1 becomes approximately the second-order functional gradient descent.  
Also, with this choice of $\Bref$, 
\algm\ can be converted to a simpler form where 
we do not have to compute the values of the guide function $\tarffx{m}$ explicitly, 
and where we have one fewer meta-parameter.  
This simpler form is shown in \algL, which has the following 
relationship to \algm.  

\begin{proposition}
\label{pr:m}
When $\Bref(p) = \lossy(p)$ that returns loss given prediction $p$, 
\algm\ with $\ssa=\ssai$ is equivalent to
\algL\ with $\ssa=1 - (1-\ssai)^m$. 
\end{proposition}
The proofs are all provided in the supplementary material. 

To simplify notation, let $\netprm=\net{\prm}{x}$, which is the model that we are updating, 
and $\netprmt=\net{\prm_t}{x}$, which is a model that was frozen when time changed from $t-1$ to $t$.  
In the stage associated with time $t$, \algL\ minimizes
\begin{equation}
  \meanOverS{ \Bdivx{\lossy}(\netprm,\netprmt) + \ssa \nabla \lossy(\netprmt)^\top \netprm} + \reg(\prm)
  \label{eq:algLobj0}
\end{equation}
approximately through mini-batch SGD. 
The second term $\ssa \nabla \lossy(\netprmt)^\top \netprm$ pushes the model $\netprm$ 
towards the direction of reducing loss, 
and the first term $\Bdivx{\lossy}(\netprm,\netprmt)$ pulls it back towards the frozen model $\netprmt$.  
With a certain family  of loss functions, \eqref{eq:algLobj0} can be further transformed 
as follows. 
\begin{proposition}
\label{pr:distill}
Let $y$ be a vector representation 
such as a $K$-dim vector representing $K$ classes.  
Assume that the gradient of the loss function can be expressed as 
\begin{equation}
  \nabla \loss(f,y) = \nabla \lossy(f) = \pp(f) - y  \label{eq:lossCond}
\end{equation}
with $\pp(f)$ not depending on $y$. Let
\begin{align}
J_t(\theta) =& \meanOverS{\Bdivx{\lossy}(\netprm, \netprmt) + \ssa \nabla \lossy(\netprmt)^\top
\netprm} 
\label{eq:algLobj} \\
J'_t(\theta) =& \meanOverS{(1-\ssa)\loss(\netprm, \pp(\netprmt)) + \ssa
\lossy(\netprm)} 
\label{eq:distillobj}              
\end{align}
Then we have
\[
J_t(\prm) = J'_t(\prm) + c_t~,
\]
where $c_t$ is independent of $\prm$.
This implies that
\[
\arg\min_\prm \left[ J_t(\prm) + R(\prm) \right]
=
\arg\min_\prm \left[ J'_t(\prm) + R(\prm) \right] .
\]
\end{proposition}
Both the cross-entropy loss and squared loss satisfy \eqref{eq:lossCond}.  
In particular, when $\lossy(f)$ is the cross-entropy loss, 
$\pp(f)$ becomes the soft max function.  
In this case, \eqref{eq:distillobj} is the {\em distillation} formula 
with the frozen model $\netprmt$ playing the role of a cumbersome source model, 
and therefore, the parameter update rule of \algL\ involving \eqref{eq:algLobj} 
becomes that of distillation.  
Thus, Algorithm~\ref{alg:AlgL} can be regarded as a generalization of
self-distillation for arbitrary loss functions.

\subsection{Convergence Analysis}

Let us define {\em $\ssa$-regularized loss} 
\begin{equation}
  \reglossx{\ssa}(\prm) := \meanOverSp{ \myloss{\net{\prm}{x}} } + \frac{1}{\ssa}\reg(\prm).  \label{eq:reglossa}
\end{equation}
The following theorem shows that 
Algorithm \ref{alg:Algm} with step-size $\ssa$ always approximately reduces the $\ssa$-regularized loss 
if $\ssa$ is appropriately set.  

\newcommand{\prmtTilde}{\tilde{\prm}}
\begin{theorem}
\label{thm:L-conv}
In the setting of \algm\ with $m=1$, assume that there exists $\beta>0$
such that $\Bdivx{h}(f,f') \geq \beta \Bdivx{\lossy}(f,f')$ for any $f$ and $f'$, and 
assume that $\ssa \in (0,\beta]$.  
Assume also that $Q_t(\prm)$ defined in \algm\ is $1/\lr$ smooth in $\prm$:
$
\|\nabla Q_t(\prm) - \nabla Q_t(\prm')\| \leq (1/\lr) \|\prm -
\prm'\| .
$

Assume that $\prm_{t+1}$ is an improvement of $\prm_t$ with respect to minimizing $Q_t$ so that 
$Q_t(\prm_{t+1}) \leq Q_t(\prmtTilde)$ ,  
where
\begin{align}
\prmtTilde = \prm_t - \lr \nabla Q_t(\prm_t) . 
\label{eq:algLupd}
\end{align}
Then we have 
\[
  \reglossx{\ssa}(\prm_{t+1}) \le \reglossx{\ssa}(\prm_t) - \frac{\ssa \lr}{2}
  \|\nabla \reglossx{\ssa}(\prm_{t})\|^2 .
\]
\end{theorem}
For \algL, we have $h(\cdot)=\lossy(\cdot)$ and thus $\beta=1$, 
leading to $\ssa\in (0,1]$.  
\eqref{eq:algLupd} is the parameter update step of \algm\ 
except that the algorithm stochastically estimates the mean over $\trnset$ 
from a mini-batch $\batch$ sampled from $\trnset$.  
Therefore, 
the theorem indicates that each stage (corresponding to $t$) of the algorithm 
approximately reduces the $\ssa$-regularized loss $\reglossx{\ssa}(\prm)$.  
In other words, while the guide function changes from stage to stage, 
a quantity that does {\em not} depend on the guide function 
goes down throughout training, namely, the $\ssa$-regularized loss $\reglossx{\ssa}$.  

Furthermore, we obtain from Theorem~\ref{thm:L-conv} that
\[
\frac{1}{T}
\sum_{t=0}^{T-1}   \|\nabla \reglossx{\ssa}(\prm_{t})\|^2 
\leq \frac{2( \reglossx{\ssa}(\prm_0)-\reglossx{\ssa}(\prm_{T}))}{\ssa \lr T} .
\]
Assuming $\reglossx{\ssa}(\prm) \ge 0$, this implies that as $\tT$ goes to infinity, 
the right-hand side goes to zero, and so \algL\ converges with 
$\nabla \reglossx{\ssa}(\prm_{\tT}) \to 0$.
Therefore, when $\tT$ is sufficiently large, $\prm_{\tT}$ finds a
stationary point of $\reglossx{\ssa}$. 

The convergence result indicates that having a regularization term $\reg(\prm)$ in the algorithm 
effectively causes minimization of the $\ssa$-regularized loss.  
However, our empirical results (shown later) indicate that 
\ours\ models are very different from standard models trained directly to
minimize the $\ssa$-regularized loss.  
For example, standard models trained with $\reglossx{0.01}$ suffers from severe 
underfitting, but \ours\ model with $\ssa$=0.01 produces high performance.  
This is because each
step of guided learning tries to find a good solution which is near
the previous solution (guidance). The complexity of each iterate is better controlled,
and hence this approach leads to better generalization performance. 
We will come back to this point in the next section.  

\section{Empirical study} 
\label{sec:empirical}
While the proposed framework is general, our empirical study places a
major focus  on \oursL\ (\algL) with the cross-entropy loss, due to its connection to distillation 
(Proposition \ref{pr:distill}).  
In particular, we set up our implementation so that one instance of \oursL\ 
coincides with 
self-distillation 
to provide empirical insight into it from a functional gradient viewpoint.  

First, with the goal of understanding the empirical behavior of the algorithm,  
we examine obtained models in reference to our theoretical findings.  
We use relatively small neural networks for this purpose.  
Next, we study the case of larger networks with consideration of practicality.  

\subsection{Implementation}
\label{sec:implement}

To implement the algorithms presented above, methods of 
parameter initialization and optimization need to be considered. 
To observe the basic behavior, our strategy in this work is to keep it as simple as possible.  

\mypara{Initial parameter $\prm_0$}
As the functions of interest are nonconvex, the outcome depends on the initial parameter $\prm_0$.  
The most natural (and simplest) choice is random parameters.  
This option is called `\iniRandom' below.  
We also considered two more options. %
One is to start from a {\em base model} obtained by regular training, 
called `\iniBase'.  
This option enables study of self-distillation.  
The other is to start from a shrunk version of the base model, 
and details of this option will be provided later.  %

\mypara{Parameter update}
To update parameter $\prm$ by descending the stochastic gradient, 
standard techniques can be used such as momentum, Rmsprop \cite{rmsprop}, 
Adam \cite{Adam15}, and so forth.  
As is the case for regular training, learning rate scheduling is beneficial.  
Among many possibilities, we chose to repeatedly use for each $t$, 
the same method that works well for regular training.  
For example, a standard method for CIFAR10 is 
to use momentum and decay the learning rate only a few times, 
and therefore we use this scheme for each stage on CIFAR10.  
That is, the learning rate is reset to the initial rate for each $t$; 
however note that $\prm$ is not reset.  
Although this is perhaps not the best strategy in terms of computational cost, 
its advantage is that at the end of each stage, we obtain ``clean'' intermediate models 
with $\prm_t$ that were optimized for intermediate goals. 
(If instead, we used one decay schedule from the beginning to the end, 
the convergence theorem still holds, but 
$\prm_t$ would be noisy when the learning rate is still high.) 
This strategy enables to study how a model changes as the guide function gradually goes ahead, 
and also relates the method to self-distillation.

Since $\prm$ is not reset when the learning rate is reset, this schedule can be regarded as 
a simplified fixed-schedule version of {\em SGD with warm restarts (SGDR)} \cite{cosine17}. 
(SGDR instead does sophisticated scheduling with cosine-shape decay and variable epochs.) 
For comparison, we test the same schedule with the standard optimization objective 
(`\baseLoop'; Algorithm \ref{alg:baseLoop}). 

\mypara{Enabling study of self-distillation}
We study classification tasks with the standard cross-entropy loss, which satisfies the condition 
of Proposition \ref{pr:distill}.  
Combined with the choice of learning rate scheduling above, 
\oursL\ with the \iniBase\ option (which initializes $\prm_0$ with a trained model) 
essentially becomes self-distillation. 
Thus, one aspect of our experiments is 
to study self-distillation from the viewpoint of functional gradient learning.   

\begin{algorithm}[t]
\begin{small}
   \caption{\baseLoop\ (simplified SGDR): 
         {\bfseries Input:} $\prm_0$, training set $\trnset$.  
         Meta-parameter: $\tT$. 
         {\bfseries Output:} $\prm_\tT$. 
   } 
   \label{alg:baseLoop}
\begin{algorithmic}
   \FOR{$t=0$ {\bfseries to} $\tT-1$}  
      \STATE $\prm_{t+1} \assign \arg\min_\prm \left[ 
               \meanOverSp{ \lossy(\net{\prm}{x}) } + \reg(\prm)
             \right]$ \\
             $~~~~~~~~~~~~~$ where $\prm$ is initialized by $\prm_t$.  
   \ENDFOR
\end{algorithmic}
\end{small}
\end{algorithm}
\subsection{Experimental setup}
\label{sec:setup}
\begin{table}[t]
\begin{center} \begin{small} \begin{tabular}{|c|r|r|r|r|}

\hline
           & \#class & \multicolumn{1}{|c|}{train} & \multicolumn{1}{|c|}{dev.} & \multicolumn{1}{|c|}{test} \\
\hline
  CIFAR10  &   10 &   49000 &  1000 & 10000 \\
  CIFAR100 &  100 &   49000 &  1000 & 10000 \\
  SVHN     &   10 &  599388 &  5000 & 26032 \\
  ImageNet & 1000 & 1271167 & 10000 & 50000 \\
\hline
\end{tabular} \end{small} \end{center}
\vskip -0.15in
\caption{\label{tab:data-stat}
  Data. 
For each dataset, we randomly split the official training set into 
a training set and a development set to use
the development set for meta-parameter tuning.  
For ImageNet, following custom, %
we used the official validation set as our `test' set.   
}
\vskip -0.1in
\end{table}
Table \ref{tab:data-stat} summarizes the data we used.  
As for network architectures, we mainly used ResNet 
\cite{Resnet16a,Resnet16b} and wide ResNet (WRN) \cite{Wresnet16}. 
Following the original work, the regularization term $\reg(\prm)$ was 
set to be $\reg(\prm)= \frac{\lam}{2}\| \prm \|^2$ 
where $\lam$ is the weight decay. 
We fixed mini-batch size to 128 and used the same learning rate decay schedule 
for all but ImageNet.  
Due to the page limit, details are described in the supplementary material.  
However, 
note that the schedule we used for all but ImageNet
is 3--4 times longer than those used in the original ResNet or WRN study
for CIFAR datasets.  
This is because we used the ``train longer'' strategy \cite{HHS17}, and 
accordingly, the base model performance visibly improved from the original work.  
This, in fact, made it harder to obtain large performance gains 
over the base models 
(not only for \ours\ but also for all other tested methods) 
as the bar was set higher.  We feel that this is more realistic 
testing than using the original shorter schedule.  

We applied the standard mean/std normalization to images 
and used the standard image augmentation.  
In particular, for ImageNet, we used the same data augmentation scheme 
as used for training the pre-trained models provided as part of TorchVision, 
since we used these models as our base model.  

The default value of $\ssa$ is 0.3. 

\begin{figure}[h]
\begin{subfigure}[b]{0.5\linewidth}
\begin{center}
\centerline{\includegraphics[width=1\linewidth]{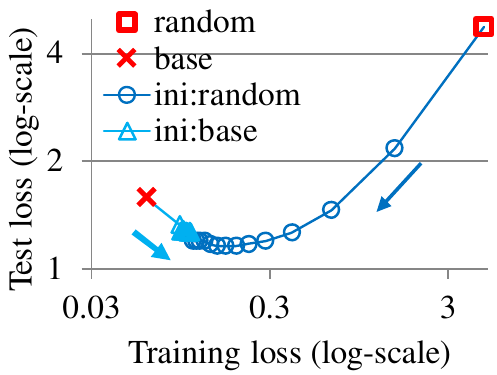}}
\vskip -0.1in
\caption{ \label{fig:c100k1-tloss-with-legend}
  \oursL
}
\end{center}
\end{subfigure}%
\begin{subfigure}[b]{0.5\linewidth}
\begin{center}
\centerline{\includegraphics[width=1\linewidth]{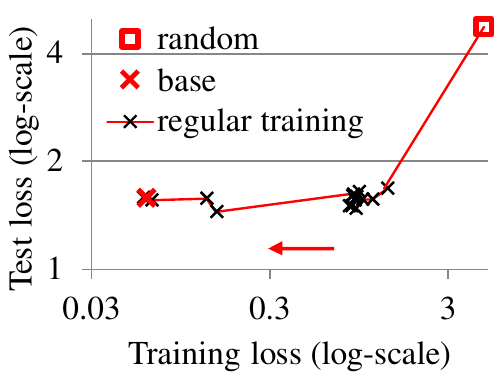}}
\vskip -0.1in
\caption{ \label{fig:c100k1-a0-tloss-with-legend}
  Regular training
}
\end{center}
\end{subfigure}%
\vskip -0.3in
\caption{ \label{fig:c100k1-tloss}
Test loss in relation to training loss. 
The arrows indicate the direction of time flow. 
CIFAR100.  ResNet-28.
} 
\vskip -0.1in
\end{figure}
\begin{figure*}
\centering
\begin{subfigure}[b]{0.2\linewidth}
\begin{center}
\centerline{\includegraphics[width=1\linewidth]{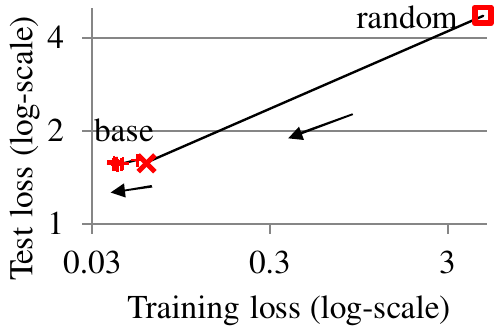}}
\vskip -0.1in
\caption{ \label{fig:c100k1-tloss-a1}
  \baseLoop\ ($\ssa$=1)
}
\end{center}
\end{subfigure}%
\begin{subfigure}[b]{0.2\linewidth}
\begin{center}
\centerline{\includegraphics[width=1\linewidth]{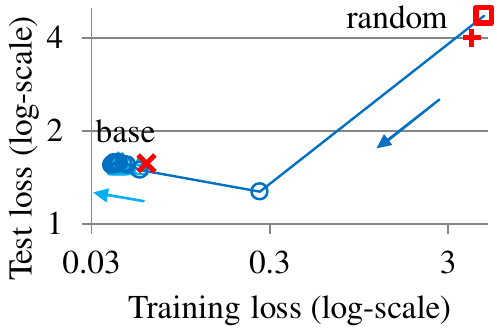}}
\vskip -0.1in
\caption{ \label{fig:c100k1-tloss-a09}
  $\ssa$=0.9
}
\end{center}
\end{subfigure}%
\begin{subfigure}[b]{0.2\linewidth}
\begin{center}
\centerline{\includegraphics[width=1\linewidth]{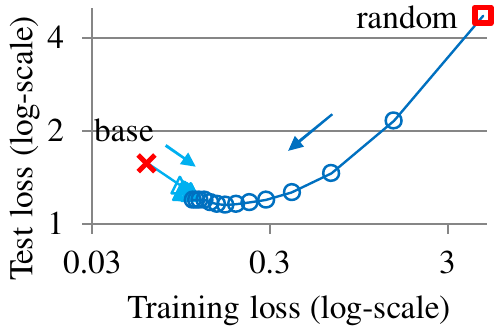}}
\vskip -0.1in
\caption{ \label{fig:c100k1-tloss-a03}
  $\ssa$=0.3
}
\end{center}
\end{subfigure}%
\begin{subfigure}[b]{0.2\linewidth}
\begin{center}
\centerline{\includegraphics[width=1\linewidth]{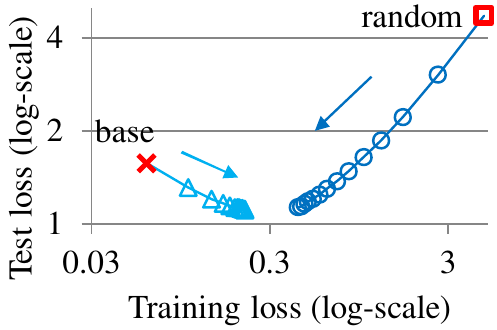}}
\vskip -0.1in
\caption{ \label{fig:c100k1-tloss-a01}
  $\ssa$=0.1
}
\end{center}
\end{subfigure}%
\begin{subfigure}[b]{0.2\linewidth}
\begin{center}
\centerline{\includegraphics[width=1\linewidth]{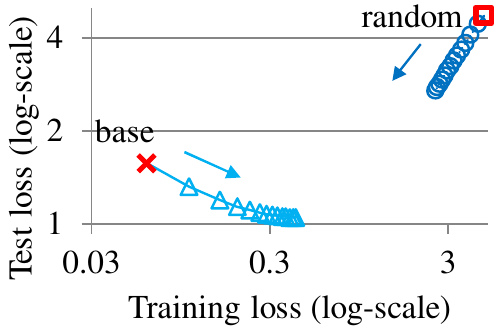}}
\vskip -0.1in
\caption{ \label{fig:c100k1-tloss-a001}
  $\ssa$=0.01
}
\end{center}
\end{subfigure}%
\vskip -0.1in
\vskip -0.25in
\caption{ \label{fig:c100k1-tloss-alpha}
Test loss %
of \iniBase (`$\triangle$') and \iniRandom(`$\circ$').  
with five values of $\ssa$ (becoming smaller from left to right), 
in relation to training loss. \oursL.  $\tT$=25. 
CIFAR100.  ResNet-28. 
As $\ssa$ becomes smaller, 
the (potential) meeting point shifts 
further away from the base model. 
The left-most figure shows \baseLoop, which is equivalent to 
$\ssa$=1. 
}
\vskip -0.1in
\end{figure*}
\subsection{Smooth path} 

We start with examining training of a relatively small network ResNet-28 (0.4M parameters) 
on CIFAR100.  
In this setting, optimization is fast, and so a relatively large $\tT$ 
(the number of stages) is feasible.  

We performed \oursL\ training with $\tT$=25 starting from random parameters (\iniRandom)
as well as starting from a base model obtained by regular training (\iniBase). 
Figure \ref{fig:c100k1-tloss-with-legend} plots test loss of these two runs 
in relation to training loss. 
Each point represents a model $\net{\prm_t}{x}$ 
at time $t=1,3,5,\cdots,25$, and the arrows
indicate the direction of time flow.  
We observe that training proceeds on a {\em smooth path}.  
\iniRandom ($\circ$), which starts from random parameters ($\square$), 
reduces both training loss and test loss.  
\iniBase ($\triangle$) starts from the base model ($\times$) and 
increases training loss, but reduces test loss.  
\iniRandom\ and \iniBase\ meet and complete one {\em smooth path} from a random state ($\square$)
to the base model ($\times$).  
\iniRandom\ goes forward on this path while \iniBase\ goes backward, 
and importantly, the path goes through the region where test loss is lower than 
that of the base model.  
The test error plotted against training loss also forms a U-shape path.  
Similar U-shape curves %
were observed across datasets and network architectures. 
The supplementary material shows 
a test error curve and 
a few more examples of test loss curves including a case of DenseNet \cite{densenet17}. 

In the middle of this path, a number of models with good generalization performance lie.  
One might wonder if regular training also forms such a path.   
Figure \ref{fig:c100k1-a0-tloss-with-legend} shows that this is not the case. 
This figure plots the loss of intermediate models 
in the course of regular training so that 
the $i$-th point represents a model after 20K$\times i$ steps of 
mini-batch SGD with the learning rate being reduced twice.   
The path of regular training from random initialization ($\square$) to the final model ($\times$) 
is rather bumpy and the test loss generally stays as high as the final outcome.   
The bumpiness is due to the fact that the learning rate is relatively high at the beginning of training.  
Comparing Figures \ref{fig:c100k1-tloss-with-legend} and \ref{fig:c100k1-a0-tloss-with-legend}, 
\ours\ training clearly takes a very different path from regular training.  

\subsection{In relation to the theory} 

\mypara{Going forward, going backward} 
It might look puzzling why \iniBase\ goes {\em backward} in the direction of {\em increasing} the training loss.  
Theorem \ref{thm:L-conv} suggests that this is the effect of the regularization term $\reg(\prm)$, 
in this case $\reg(\prm)= \frac{\lam}{2}\|\prm\|^2$ with weight decay $\lam$. 
The theory indicates that for $\ssa \in (0,1]$, the $\ssa$-regularized loss 
\[
  \reglossa = \meanOverS{ \lossy(\net{\prm}{x}) } + \reg(\prm)/\ssa
\]  
goes down and eventually converges as \oursL\ proceeds.  By contrast, 
The base model is a result of minimizing 
\[
  \meanOverS{ \lossy(\net{\prm}{x}) } + \reg(\prm). %
\]
As we always set $\ssa<1$ (0.3 in this case), i.e., $1/\ssa>1$, 
\oursL\ prefers smaller parameters than the base model does.  
Consequently, 
when \oursL\ (with small $\ssa$) starts from the base model (which has low training loss and high $\reg(\prm)$ ), 
\oursL\ is likely to reduce $\reg(\prm)/\ssa$ at the expense of increasing loss (going backward).  
When \oursL\ starts from random parameters, whose training loss is high, 
\oursL\ is likely to reduce loss (going forward) 
at the expense of increasing $\reg(\prm)/\ssa$. 

\mypara{Effects of changing $\ssa$} 
With \oursL, the guide function $\tarff$ satisfies
\begin{align*}
  \tarff &\approx \netprmt - \ssa (\hessian{\lossy(\netprmt)})^{-1} \nabla \lossy(\netprmt), 
\end{align*}
thus, $\ssa$ serves as a step-size of functional gradient descent for {\em reducing loss}.  
The effects of changing $\ssa$ are shown in Figure \ref{fig:c100k1-tloss-alpha} 
with $\tT$ fixed to 25.  
The left-most graph is \baseLoop, which is equivalent to \oursL\ with $\ssa$=1 in 
this implementation. 
There are three things to note.  
First, with a very small step-size $\ssa$=0.01 (the right most), \iniRandom\ cannot reach far from the random state 
for $\tT$=25. 
This is a straightforward effect of a small step size.   
Second, 
as step-size $\ssa$ becomes smaller (from left to right), 
the (potential) meeting/convergence point shifts further away from the base model; 
the convergence point of $\|\prm_t\|^2$ also shifts away from the base model 
and decreases (supplementary material). 
This is the effect of larger $\reg(\prm)/\ssa$ for smaller $\ssa$.  
Finally, with a large step-size (0.9 and 1), the curve flattens and 
it no longer goes through the high-performance regions {\em slowly} or {\em smoothly}, 
and the benefit diminishes/vanishes. 

\begin{figure}[h]
\vskip -0.1in
\begin{subfigure}[b]{0.5\linewidth}
\begin{center}
\centerline{\includegraphics[width=1\linewidth]{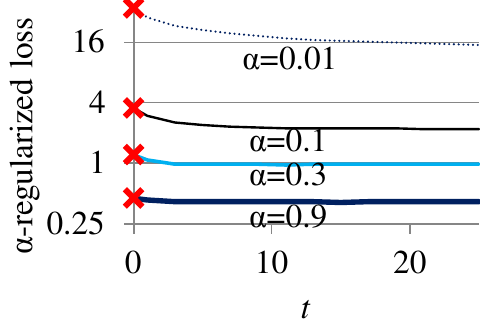}}
\vskip -0.1in
\caption{ \label{fig:c100k1-ajl2l-alpha}
}
\end{center}
\end{subfigure}%
\begin{subfigure}[b]{0.5\linewidth}
\begin{center}
\centerline{\includegraphics[width=1\linewidth]{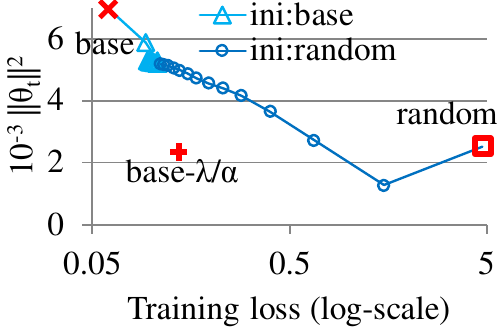}}
\vskip -0.1in
\caption{ \label{fig:c100k1-l2-baselam-a03}
}
\end{center}
\end{subfigure}%
\vskip -0.3in
\caption{ \label{fig:c100k1-ajl2l-baselam-alpha}
(a) $\ssa$-regularized loss $\reglossa$ in relation to time $t$. \oursL\ \iniBase. 
(b) $\|\prm_t\|^2$ and training loss of \baseLam\ in comparison with \oursL.  $\ssa$=0.3. 
CIFAR100.  ResNet-28. 
}
\end{figure}

\mypara{$\ssa$-regularized loss $\reglossa$} 
Figure \ref{fig:c100k1-ajl2l-alpha} confirms that, as suggested by the theory, 
 $\reglossa$ goes down and almost converges as training proceeds.  
This fact motivates examining standard models trained with this $\reglossa$ objective, 
which we call \baseLam\ models.  
We found that \baseLam\ models do not perform as well as \oursL\ at all.  In particular, 
with a very small $\ssa$=0.01, which 100 times tightens regularization, 
test error of \baseLam\ drastically degrades due to underfitting; 
in contrast, \iniBase\ with $\ssa$=0.01 performs well. 
Moreover,  
\baseLam\ models are very different from \oursL\ models with corresponding $\ssa$
even with a moderate $\ssa$.  
For example, 
Figure \ref{fig:c100k1-l2-baselam-a03} plots the parameter size $\|\prm_t\|^2$ 
in relation to training loss for $\ssa$=0.3. \baseLam\ is clearly far away from 
where \iniBase\ and \iniRandom\ converge to.  
\vskip -0.1in
\mypara{Benefit of guiding} 
This fact illustrates the merit of guided learning (including self-distillation). 
\ours\ (indirectly and locally) minimizes the $\ssa$-regularized loss $\reglossa$, but it does this 
against the restraining force of {\em pulling the model back} to the current model.  This serves as 
a form of regularization.  Without such a force, training for, say, $\reglossx{0.01}$ 
would make a big jump to rapidly reduce the parameter size and end up with 
a radical solution that suffers severely from underfitting.  This is what happens with \baseLam. 
By contrast, guided learning finds a more moderate solution with good generalization performance, 
and this is the benefit of extra regularization (in the form of pulling back) provided through the 
guide function.  
The regularization effect of distillation has been mentioned \cite{distill14}, 
and our framework formalizes the notion through the functional gradient learning viewpoint. 

\subsection{With smaller networks}

\newcommand{\imp}{\em} %
\begin{table}[b]
\vskip -0.2in
\begin{center} \begin{small} \begin{tabular}{c|c|c|c|c|c|c|}
\cline{2-7}
  &\multicolumn{2}{|c|}{}& C10 & C100 & \multicolumn{2}{|c|}{SVHN} \\
\cline{2-7}
1 & \multirow{4}{*}{baselines} 
  & \baseInTable  & 6.42 & 30.90 & 1.86 & 1.64 \\ 
\cline{3-7}  
2 && \baseLam     & 6.60 & 30.24 & 1.78 & 1.67     \\ %
3 && \baseLoop    & 6.20 & 30.09 & 1.93 &{\bf 1.53}\\
4 && label smooth & 6.66 & 30.52 & 1.71 & 1.60     \\
\cline{2-7}
5 & \multirow{2}{*}{\oursL}   
   & \iniRandom &     5.91 &{\bf 28.83} &     1.71 &{\bf 1.53} \\
6 && \iniBase   &{\bf 5.75}&     29.12  &{\bf 1.65}&     1.56  \\
\cline{2-7}
\end{tabular} \end{small} \end{center}
\vskip -0.125in
\caption{\label{tab:c10-c100-svhn}
  Test error (\%).  Median of 3 runs.
  Resnet-28 (0.4M parameters) for CIFAR10/100, 
  and WRN-16-4 (2.7M parameters) for SVHN.  
  Two numbers for SVHN are without and with dropout. 
  \baseLam: weight decay $\lam/\ssa$.  
  \baseLoop: Algorithm \ref{alg:baseLoop}.  
}
\vskip -0.1in 
\end{table}

Now we review the test error results of using relatively small networks 
in Table \ref{tab:c10-c100-svhn}.  
$\tT$ for \oursL\ and \baseLoop\ was fixed to 25 on CIFAR10/100 
and 15 on SVHN.  
Step-size $\ssa$ was fixed to 0.3 for \iniRandom\ and 
chosen from $\{$ 0.01,0.03 $\}$ for \iniBase.  
\oursL\ is consistently better than the base model (Row 1)
and generally better than the three baseline methods (Row 2--4).  
The \baseLam\ results (Row 2) were obtained by $\ssa$=0.3, and they are generally 
not much different from the base model.  
\baseLoop\ (Row 3) generally makes small improvement over the base model, 
but it generally falls short of \oursL.  
A common technique, label smoothing (Row 4) \cite{inception16}, 
`softens' labels by taking a small amount of 
probability from the correct class and distributing it equally to the incorrect classes.  
It 
generally worked well, but the improvements were small. 
That is, 
the three baseline methods produced performance gains to some extent, but 
their gains are relatively small, and they are not as consistent as \oursL\ across datasets.  

\mypara{\iniRandom} 

In these experiments, 
\iniRandom\ performed as well as \iniBase.  This fact cannot be explained from 
the knowledge-transfer viewpoint of distillation, but it can be explained from our 
functional gradient learning viewpoint, as in the previous section.  

\begin{figure}[t]
\begin{subfigure}[b]{0.5\linewidth}
\begin{center}
\centerline{\includegraphics[width=1\linewidth]{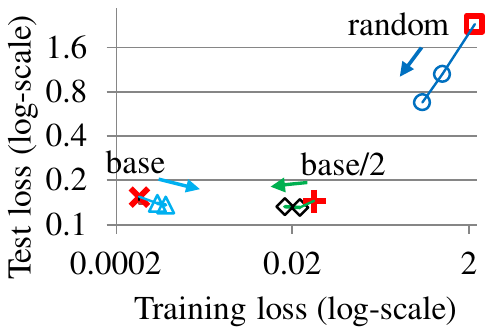}}
\vskip -0.1in
\caption{ \label{fig:c10k10-tloss-with-fcs}
  CIFAR10
}
\end{center}
\end{subfigure}%
\begin{subfigure}[b]{0.5\linewidth}
\begin{center}
\centerline{\includegraphics[width=1\linewidth]{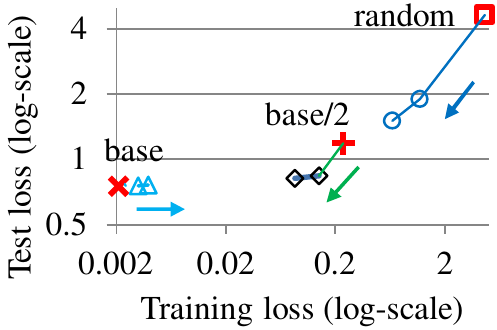}}
\vskip -0.1in
\caption{ \label{fig:c100k10-tloss-with-fcs}
  CIFAR100
}
\end{center}
\end{subfigure}%
\vskip -0.25in
\caption{ \label{fig:c10c100k10-tloss-with-fcs}
  Test loss in relation to training loss. 
  WRN-28-10 on CIFAR10 and CIFAR100. \oursL. 
  \iniBase/2 (`$\diamond$') fills the gap between \iniRandom\ (`$\circ$') and \iniBase\ (`$\triangle$'). 
}
\end{figure}
\begin{table}[t]
\begin{center} \begin{small} \begin{tabular}{c|c|c|c|}
\cline{2-4}
&     & CIFAR10 & CIFAR100 \\
\cline{2-4}
1&\baseInTable& 3.82 & 18.55 \\ 
\cline{3-4}  
2&\baseLam    & 3.70 & 27.89 \\
3&\baseLoop  & 3.70 & 18.91 \\  %
4&lab smooth & 4.13 & 19.44 \\  %
\cline{2-4}
5& \ourssq  &{\bf 3.46}& 18.14 \\ %
6& \oursL   & 3.63 &{\bf 17.95}\\ %
\cline{2-4}
\end{tabular} \end{small} \end{center}
\vskip -0.15in
\caption{\label{tab:c10-c100-k10}
Test error (\%) results on CIFAR10 and CIFAR100.  
WRN-28-10 (36.5M parameters) without dropout.  
Median of 3 runs. 
}
\vskip -0.1in
\end{table}
\subsection{With larger networks} 
The neural networks and the size of images (32$\times$32) used above are relatively small.  
We now consider computationally more expensive cases. %

\mypara{Parameter shrinking}
\iniRandom, the most natural option from the functional gradient learning viewpoint, 
unfortunately, turned out to be too costly in this large-network situation.  
Moreover, in this setting, 
it is useful to have an option of starting 
somewhere between the two end points (`random' and `base') 
since that is where good models tend to lie according to our study with small networks.  
Therefore, we experimented with `rewinding' a base model 
by shrinking its weights and bias of the last fully-connected linear layer 
by dividing them with $\fcsmeta > 1$ (a meta-parameter).  
We use these partially-shrunk parameters as the initial parameter $\prm_0$ for 
\ours.  Since doing so shrinks the model output $\net{\prm_0}{x}$ 
by the factor of $\fcsmeta$, this is closely related to {\em temperature scaling}, 
for {\em distillation} \cite{distill14} and post-training calibration \cite{GPSW17}.  
Parameter shrinking is, however, simpler than temperature scaling of distillation, 
which scales logits of both models, 
and fits well in our framework. 

Figure \ref{fig:c10c100k10-tloss-with-fcs} shows 
training loss (the $x$-axis) and test loss (the $y$-axis) 
obtained when parameter shrinking is applied to WRN-28-10 
on CIFAR10 and CIFAR100.  
By shrinking with $\fcsmeta$=2, the loss values of the base model change 
from `\baseInFig' ($\times$)
to `\baseInFig/2' ($+$). 
The location of \baseInFig/2 is roughly the midpoint of two end points 
`\baseInFig' and `random'. 
\iniBase/2 ($\diamond$), which starts from the shrunk model, explores the space
neither \iniRandom\ nor \iniBase\ can reach in a few stages. 
\mypara{Larger ResNets on CIFAR10 and CIFAR100}

Table \ref{tab:c10-c100-k10} shows test error of \iniBase/2 
using WRN-28-10 on CIFAR10/100. 
$\tT$ was fixed to 1.  
Compared with the base model, 
both \ours1 and 2 consistently improved performance, 
while the baseline methods mostly failed to make improvements.  
\ours1 and 2 produced similar performances. %
This is the best WRN non-ensemble results on CIFAR10/100 
among the self-distillation related studies that we are aware of.  
\mypara{ImageNet}

To test further scale-up, 
we experimented with 
ResNet-50 (25.6M parameters) and WRN50-2 (68.9M parameters)
on the ILSVRC-2012 ImageNet dataset. 
As ImageNet training is resource-consuming, we only tested 
selected configurations, which are \oursL\ with \iniBase\ and \iniBase/2 options. 
In these experiments, $\ssa$ was set to 0.5, but partial results suggested that 0.3  
works well too.  
We used models pre-trained on ImageNet provided as part of TorchVision\footnote{
  https://pytorch.org/docs/stable/torchvision/models.html
}
as the base models.  
Table \ref{tab:imgnet} shows that 
\oursL\ consistently improves error rates over the base model.  
The best-performing \iniBase/2 achieved lower error rates than 
a twice deeper counterpart of each network, 
ResNet-101 for ResNet-50 and WRN-101-2 for WRN-50-2
trained in a standard way (Rows 11--12).  
Thus, we confirmed that \oursL\ scales up and brings performance gains on ImageNet.  
To our knowledge, this is one of the largest-scale 
ImageNet experiments among the self-distillation related studies.  
\begin{table}[h]
\begin{center} \begin{small} \begin{tabular}{c|c|c|cc|cc|}
\cline{2-7} %
&\multicolumn{2}{|c|}{methods}
                   & \multicolumn{2}{|c|}{Resnet-50}
                   & \multicolumn{2}{|c|}{WRN50-2} \\                  
\cline{2-7} \noalign{\vskip\doublerulesep \vskip-\arrayrulewidth} \cline{2-7}
1&\multicolumn{2}{|c|}{\baseInTable} & 23.87 & 7.14 & 21.53 & 5.91 \\ %
\cline{2-7}
2&\multirow{3}{*}{\baseLoop}
           & $t$=1 & 23.73 & 6.95 & 21.99 & 6.11 \\
3&         & $t$=2 & 23.50 & 6.93 & \multicolumn{2}{|c|}{--} \\
4&         & $t$=3 & 23.36 & 6.78 & \multicolumn{2}{|c|}{--} \\
\cline{2-7} \noalign{\vskip\doublerulesep \vskip-\arrayrulewidth} \cline{2-7}
5&\multirow{3}{*}{\iniBase}
           & $t$=1 & 22.79 & 6.43 & 21.17 & 5.65 \\ %
6&         & $t$=2 & 22.49 & 6.27 & \multicolumn{2}{|c|}{--} \\
7&         & $t$=3 & 22.31 & 6.28 &       &      \\    
\cline{2-7}
\cline{2-7}
8&\multirow{3}{*}{\iniBase/2}          
           & $t$=1 &     22.50  &     6.25 &{\bf 20.69}&{\bf 5.35}\\
9&         & $t$=2 &     22.31  &     6.18 & \multicolumn{2}{|c|}{--}  \\
10&         & $t$=3 &{\bf 22.08} &{\bf 6.10}& \multicolumn{2}{|c|}{--}  \\
\cline{2-7} \noalign{\vskip\doublerulesep \vskip-\arrayrulewidth} \cline{2-7}
11&\multicolumn{2}{|c|}{Resnet-101$\dagger$} & 22.63 & 6.44 & \multicolumn{2}{|c|}{--} \\
12&\multicolumn{2}{|c|}{WRN-101-2$\dagger$}  &  \multicolumn{2}{|c|}{--} & 21.16 & 5.72 \\
\cline{2-7}
\end{tabular} \end{small} \end{center}
\vskip -0.15in
\caption{\label{tab:imgnet}
ImageNet 224$\times$224 single-crop results on the validation set. 
\oursL. 
top-1 and top-5 errors (\%). \\
$\dagger$
The ResNet-101 and WRN-101-2 performances are from the description of 
  the pre-trained torchvision models.  
}
\vskip -0.1in
\end{table}

\mypara{Additional experiments on text}
Finally, the experiments in this section used image data.  Additional experiments 
using text data are presented in the supplementary material.

\section{Discussion}

\mypara{Guided exploration of landscape}
\ours\ is an informed/guided exploration of the loss landscape, 
where the guidance is successively given as interim goals set in the neighborhood of the model at the time, 
and such guidance is provided by gradient descent in a function space.  
Another view of this process is an accumulation of successive greedy optimization.  
Instead of searching the entire space for the ultimate goal of loss minimization at once, 
guided learning proceeds with repeating a local search, which limits the space to be searched
and leads to better generalization.  
Its benefit is analogous to that of $\varepsilon$-boosting. 

\mypara{\ourssq}
\oursL\ uses the second-order information of loss in the functional gradient step 
for generating the guide function, and \ourssq\ does not.  
\oursL's update rule is equivalent to that of distillation, and 
\ourssq's is not. \ourssq\ also differs from the logit least square fitting version of distillation.  
In our experiments (though limited due to our focus on self-distillation study), 
\ourssq\ performed as well as \oursL. %
If this is a general trend, this indicates that 
inclusion of the second-order information is not particularly helpful.  
If so, this could be because the second-order information is useful for accelerating optimization, 
but we would like to proceed slowly to obtain better generalization performance.  
This motivates further investigation of \ourssq\ as well as other instantiations of the framework.  

\mypara{Computational cost}
From a practical viewpoint, a shortcoming of the particular setup tested here 
(but not the general framework of \ours) 
is computational cost.  
Since we used the same learning rate scheduling as regular training in each stage, 
\ours\ training with $\tT$ stages took more than $\tT$ times longer than 
regular training.  
It is conceivable that training in each stage can be shortened without hurting 
performance since optimization should be easier as a results of aiming at a nearby goal. 
Schemes that decay the learning rate throughout the training without restarts 
or 
hybrid approaches
might also be beneficial 
for reducing computation.  Note that Theorem \ref{thm:L-conv} does not require each 
stage to be performed to the optimum.  
On the other hand, testing (i.e., making predictions) of the models trained with \ours\ 
only requires the same cost as regular models.  
As shown in the ImageNet experiments, a model trained with \ours\ 
could perform better than a much larger (and so slow-to-predict) model; 
in that case, \ours\ can save the overall computational cost 
since the cost for making predictions can be significant 
for practical purposes.

\mypara{Relation to other methods} 
The proposed method seeks to improve generalization performances in a principled way 
that limits the searched parameter space.  The relation to existing methods for similar purposes 
is at least two-fold.  First, we view that this work gives theoretical insight into related methods 
such as self-distillation and label smoothing, which we hope can be used to improve them.  
Second, methods derived from this framework can be used {\em with} existing techniques 
that are based on different principles (e.g., weight decay and dropout) for {\em further} improvements.  

\mypara{Distillation} 
Due to the connection discussed above, 
our theoretical and empirical analyses of \oursL\ provide a new functional gradient view of distillation.  
Here we discuss a few self-distillation studies from this new viewpoint. 
\cite{Ban18} showed that iterative self-distillation improves performance over the base model. 
They set $\ssa$ to 0 (in our terminology) and 
reported that there were no performance gains on CIFAR10. 
According to our theory, 
when $\ssa$ goes to 0, the quantity reduced throughout the process is 
not the $\ssa$-regularized loss but merely $\reg(\prm)$.  
Such an extreme setting might be risky.  
In {\em deep mutual learning} \cite{deepMutual18}, 
multiple models are simultaneously trained by reducing loss and aligning 
each other's model output.  They were surprised by the fact that 
`no prior powerful teacher' was necessary. 
This fact can be explained by our functional gradient view by 
relating their approach to our \iniRandom.  
Finally, the regularization effect of distillation has been noticed \cite{distill14}. 
Our framework formalized the notion through the functional gradient learning viewpoint.

\section{Conclusion}

This paper introduces a new framework for guided learning of nonconvex models
through successive functional gradient optimization. A convergence
analysis is established for the proposed approach, and it is shown that
our framework generalizes the popular self-distillation method. Since
the guided learning approach learns nonconvex models in
restricted search spaces, we obtain better generalization performance
than standard training techniques. 

\section*{Acknowledgements}

We thank Professor Cun-Hui Zhang for his support of this research.  

\bibliography{breg}
\bibliographystyle{icml2020}

\clearpage
\onecolumn

\appendix

\section{Proofs}

In the proofs, 
we use abbreviated notation by dropping $x$ and $y$ and making $\prm$ a subscript, e.g., 
we write $\netprm$ for $\netprmLong$.  

\subsection{Proof of Proposition \ref{pr:m}}

\paragraph{Proposition \ref{pr:m}}
{\em 
  When $\Bref(p) = \lossy(p)$ that returns loss given prediction $p$, 
  \algm\ with $\ssa=\ssai$ is equivalent to
  \algL\ with $\ssa=1 - (1-\ssai)^m$. 
}
\begin{proof}
From \algm\ with $\ssa=\ssai$, we have
\begin{align}
   \tarffx{i} = \arg\min_\tar \left[ 
                \Bdiv(\tar,\tarffx{i-1}) + \ssai \nabla \lossy(\tarffx{i-1})^\top \tar
   \right ] . \label{eq:tarffxi}
\end{align}
From $\Bref(\cdot)=\lossy(\cdot)$ and \eqref{eq:tarffxi}, we obtain 
\begin{align*}
\nabla\lossy(\tarffx{i}) &= \nabla\lossy(\tarffx{i-1}) - \ssai\nabla\lossy(\tarffx{i-1}) = (1-\ssai)\nabla\lossy(\tarffx{i-1})
~~\mbox{ for }i=1,\cdots,m .
\end{align*}
Since $\tarffx{0}=\netprmt$, we have 
\begin{align*}
\nabla\lossy(\tarffx{m}) &= (1-\ssai)^m \nabla\lossy(\tarffx{0})
= (1-\ssai)^m \nabla\lossy(\netprmt) , 
\end{align*}
which implies %
\begin{align*}
\nabla_{\netprm} \left[ \Bdiv(\netprm,\tarffx{m}) \right ] 
&=  \nabla\lossy(\netprm) - \nabla\lossy(\tarffx{m})
 =  \nabla\lossy(\netprm) - (1-\ssai)^m \nabla\lossy(\netprmt) \\
&=  \nabla_{\netprm} \left[ \Bdivx{\lossy}(\netprm, \netprmt) + ( 1 - (1-\ssai)^m) \nabla \lossy(\netprmt)^\top \netprm \right]
\end{align*}
and therefore, 
$
\nabla_\prm \left[ \Bdiv(\netprm,\tarffx{m}) \right ] = 
\nabla_\prm \left[ \Bdivx{\lossy}(\netprm, \netprmt) + ( 1 - (1-\ssai)^m) \nabla \lossy(\netprmt)^\top \netprm \right]
$.  The rest is trivial. 
\end{proof}

\subsection{Proof of Proposition \ref{pr:distill}}
\paragraph{Proposition \ref{pr:distill}}
{\em 
Let $y$ be a vector representation 
such as a $K$-dim vector representing $K$ classes.  
Assume that the gradient of the loss function can be expressed as 
\[
  \nabla \loss(f,y) = \nabla \lossy(f) = \pp(f) - y  
\]
with $\pp(f)$ not depending on $y$. Let
\begin{align*}
J_t(\theta) =& \meanOverS{\Bdivx{\lossy}(\netprm, \netprmt) + \ssa \nabla \lossy(\netprmt)^\top
\netprm} \\
J'_t(\theta) =& \meanOverS{(1-\ssa)\loss(\netprm, \pp(\netprmt)) + \ssa
\lossy(\netprm)} 
\end{align*}
Then we have
\[
J_t(\prm) = J'_t(\prm) + c_t,
\]
where $c_t$ is independent of $\prm$.
This implies that
\[
\arg\min_\prm \left[ J_t(\prm) + R(\prm) \right]
=
\arg\min_\prm \left[ J'_t(\prm) + R(\prm) \right] .
\]
} %
\begin{proof}
\begin{align*}
\nabla_{\netprm} \left[ \Bdivx{\lossy}(\netprm, \netprmt) + \ssa \nabla \lossy(\netprmt)^\top \netprm \right ]
&= \nabla\lossy(\netprm) - (1-\ssa)\nabla\lossy(\netprmt) \\
&= (\pp(\netprm)-y) - (1-\ssa)(\pp(\netprmt)-y) \\
&= (1-\ssa)(\pp(\netprm) - \pp(\netprmt)) + \ssa( \pp(\netprm) - y ) \\
&= \nabla_{\netprm} \left[ (1-\ssa)\loss(\netprm, \pp(\netprmt)) +
  \ssa\lossy(\netprm) \right ] .
\end{align*}  
This implies that $\nabla J_t(\prm) = \nabla
J'_t(\prm)$. Therefore $J_t(\prm) - J'_t(\prm)$ is independent of $\prm$. 
\end{proof}

\subsection{Proof of Theorem \ref{thm:L-conv}}

\paragraph{Theorem \ref{thm:L-conv}} 
{\em 
In the setting of \algm\ with $m=1$, assume that there exists $\beta>0$
such that $\Bdivx{h}(f,f') \geq \beta \Bdivx{\lossy}(f,f')$ for any $f$ and $f'$, and 
assume that $\ssa \in (0,\beta]$.  
Assume also that $Q_t(\prm)$ defined in \algm\ is $1/\lr$ smooth in $\prm$:
\[
\|\nabla Q_t(\prm) - \nabla Q_t(\prm')\| \leq (1/\lr) \|\prm - \prm'\| .
\]

Assume that $\prm_{t+1}$ is an improvement of $\prm_t$ with respect to minimizing $Q_t$ so that 
\[
 Q_t(\prm_{t+1}) \leq Q_t(\prmtTilde) , 
\]
where
\begin{align*}
\prmtTilde = \prm_t - \lr \nabla Q_t(\prm_t) . 
\end{align*}
Then we have 
\[
  \reglossx{\ssa}(\prm_{t+1}) \le \reglossx{\ssa}(\prm_t) - \frac{\ssa \lr}{2}
  \|\nabla \reglossx{\ssa}(\prm_{t})\|^2 .
\]
}
\begin{proof}

We first define $\tilde{Q}_t(\prm)$ as follows:
\[
\tilde{Q}_t(\prm) := \meanOverS{\Bdivx{h}(\netprm,\netprmt) + \ssa \nabla
  \lossy(\netprmt)^\top \netprm } + R(\prm) .
\]
We can check that $Q_t(\prm) - \tilde{Q}_t(\prm)$ is independent of
$\prm$. Therefore optimizing $\prm$ with respect to $Q_t(\prm)$ is the
same as optimizing $\prm$ with respect to $\tilde{Q}_t(\prm)$, and
$\nabla Q_t(\prm)=\nabla \tilde{Q}_t(\prm)$.

The smoothness assumption implies that
\[
\tilde{Q}_t(\prm- \Delta \prm) \leq \tilde{Q}_t(\prm) - \nabla {Q}_t(\prm)^\top \Delta
\prm + \frac{1}{2\eta } \|\Delta \prm\|^2 .
\]
Therefore
\begin{align*}
\tilde{Q}_t(\prm_{t+1}) \leq& \tilde{Q}_t(\prmtTilde)
= \tilde{Q}_t(\prm_t - \lr \nabla Q_t(\prm_t)) \\
\leq & \tilde{Q}_t(\prm_t) - \lr \|\nabla Q_t(\prm_t)\|^2 +
       \frac{1}{2\lr} \| \lr \nabla Q_t(\prm_t)\|^2 \\
=& \tilde{Q}_t(\prm_t) - \frac{\lr}{2} \|\nabla Q_t(\prm_t)\|^2 .
\end{align*}
Note also that
\begin{align*}
\tilde{Q}_t(\prm_{t+1}) - \tilde{Q}_t(\prm_t) \geq&
\meanOverS{\beta \Bdivx \lossy(f_{\prm_{t+1}},f_{\prm_t}) + \ssa \nabla \lossy(f_{\prm_t})^\top
                                 (f_{\prm_{t+1}} -f_{\prm_t}) }
+ [R(\prm_{t+1}) - R(\prm_t)]\\
=&
\meanOverS{ 
(\beta-\ssa) \Bdivx \lossy(f_{\prm_{t+1}},f_{\prm_t}) + \ssa
   \lossy(f_{\prm_{t+1}}) - \ssa \lossy(f_{\prm_t})}
+ [R(\prm_{t+1}) - R(\prm_t)]\\
\geq&
\meanOverS{ 
\ssa   \lossy(f_{\prm_{t+1}}) - \ssa \lossy(f_{\prm_t})}
+ [R(\prm_{t+1}) - R(\prm_t)]\\
=& \ssa \reglossx{\ssa}(\prm_{t+1}) - \ssa \reglossx{\ssa}(\prm_{t}) .
\end{align*}
The second inequality is due to the non-negativity of the Bregman divergence.  

By combining the two inequalities, we obtain
\[
\ssa \reglossx{\ssa}(\prm_{t+1}) \leq \ssa \reglossx{\ssa}(\prm_{t}) -
\frac{\lr}{2} \|\nabla Q_t(\prm_t)\|^2 .
\]
Now, observe that $\nabla Q_t(\prm_t) = \nabla
\tilde{Q}_t(\prm_t) =\ssa \nabla
\reglossx{\ssa}(\prm_t)$, and we obtain the desired bound.
\end{proof}

\section{On the Empirical Study} 

In this section, we first 
provide experimental details and additional figures regarding the experiments 
reported in the main paper, 
and then we report additional experiments using text data.  
Our code is provided at a repository under {\tt github.com/riejohnson}.  

\subsection{Details of the experiments in the main paper}

\subsubsection{CIFAR10, CIFAR100, and SVHN}

This section describes the experimental details of all but the ImageNet experiments.  

The mini-batch size was set to 128.  We used momentum 0.9.  
The following learning rate scheduling was used: 
200K steps with $\lr$, 40K steps with $0.1\lr$, and 40K steps with $0.01\lr$.  
The initial learning rate $\lr$ was set to 0.1 on CIFAR10/100 
and 0.01 on SVHN, following \cite{Wresnet16}. 
The weight decay $\lam$ was 0.0001 except that 
it was 0.0005 for (CIFAR100, WRN-28-10) and SVHN.  

We used the standard mean/std normalization on all 
and the standard shift and horizontal flip image augmentation on CIFAR10/100.  

We report the median of three runs with three random seeds.  
The meta-parameters were chosen based on the performance on the development set.  
All the results were obtained by using 
only the `train' portion (shown in Table \ref{tab:data-stat} of the main paper) 
of the official training set as training data.  

For label smoothing, the amount of probability taken away from the true class was chosen 
from $\{0.1, 0.2, 0.3, 0.4\}$.  

To obtain the results reported in Table \ref{tab:c10-c100-svhn} (with smaller networks), 
$\tT$ was fixed to 25 for CIFAR10/100, and 15 for SVHN.  
$\ssa$ for \iniRandom\ was fixed to 0.3. 
For \iniBase, we chose $\ssa$ from $\{ 0.3, 0.01 \}$.  
We excluded $\ssa=0.01$ for \iniRandom, as it takes too long. 
When dropout was applied in the SVHN experiments, 
the dropout rate was set to 0.4, following \cite{Wresnet16}. 
To obtain the results reported in Table \ref{tab:c10-c100-k10} (with larger networks), 
$\tT$ was fixed to 1.  
For \oursL, $\ssa$ was chosen from $\{ 0.3, 0.01 \}$. 
For \ourssq, $\ssa$ was fixed to 0.3, and $m$ 
(the number of functional gradient steps) was chosen from 
$\{ 1, 2, 5 \}$.  
On CIFAR datasets, the choice of $\ssa$ or $m$ did not make much difference, 
and the chosen values tended to vary among the random seeds.  
On SVHN, $\ssa$=0.01 tended to be better when no dropout was used, 
and 0.3 was better when dropout was used.  

To perform random initialization of the parameter for \iniRandom\ and 
the baseline methods, we used  
Kaiming normal initialization \cite{HZRS15}, following the previous work. 

\subsubsection{ImageNet}

Each stage of the training for ImageNet followed the code used for training the 
pre-trained models provided as part of TorchVision: 
{\tt https://github.com/pytorch/examples/blob/master/imagenet/main.py}.  
That is, for both ResNet-50 and WRN-50-2, 
the learning rate was set to 
$\lr$, $0.1\lr$, and $0.01\lr$ for 30 epochs each, 
i.e., 90 epochs in total, and the initial rate $\lr$ was set to 0.1. 
The mini-batch size was set to 256, and the weight decay was set to 0.0001. 
The momentum was 0.9.  
$\ssa$ was fixed to 0.5. 
We used two GPUs for ResNet-50 and four GPUs for WRN-50-2. 

We used the standard mean/std normalization 
and the standard image augmentation for ImageNet -- 
random resizing, cropping and horizontal flip, 
which is the same data augmentation scheme 
as used for training the pre-trained models provided as part of TorchVision. 

\clearpage
\subsection{Additional figures}

\begin{figure}[ht]
\begin{center}
\centerline{\includegraphics[width=0.25\linewidth]{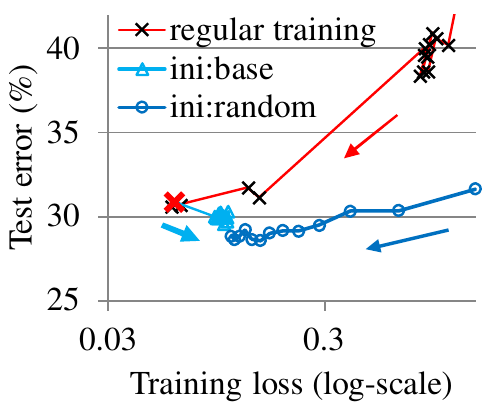}}
\vskip -0.15in
\caption{ \label{fig:c100k1-joint-terr-with-legend}
  Test error (\%) in relation to training loss. 
  The arrows indicate the direction of time flow.   
  \oursL. CIFAR100.  ResNet-28.
} 
\end{center}
\vskip 0.2in
\begin{center}
\begin{subfigure}[b]{0.25\linewidth}
\begin{center}
\centerline{\includegraphics[width=1\linewidth]{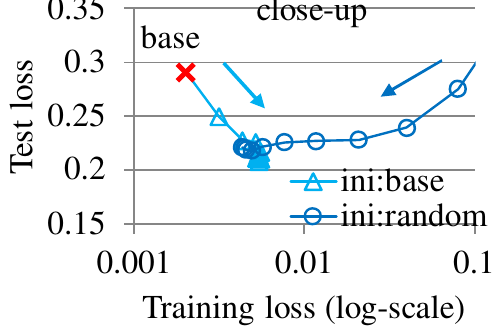}}
\label{fig:c10k1-tloss-zoom}
\vskip -0.1in
\caption{CIFAR10, ResNet-28}
\end{center}
\end{subfigure}%
\quad
\begin{subfigure}[b]{0.25\linewidth}
\begin{center}
\centerline{\includegraphics[width=1\linewidth]{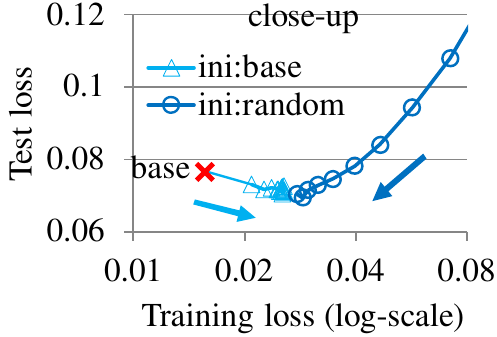}}
\label{fig:svhn-tloss-zoom}
\vskip -0.1in
\caption{SVHN, WRN-16-4.}
\end{center}
\end{subfigure}%
\quad
\begin{subfigure}[b]{0.25\linewidth}
\begin{center}
\centerline{\includegraphics[width=1\linewidth]{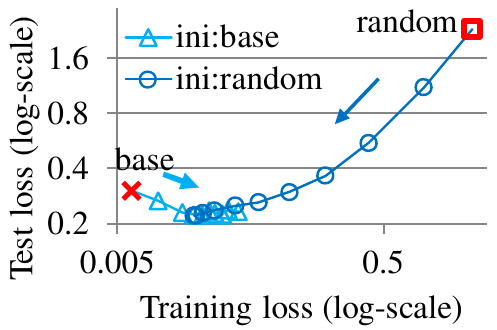}}
\label{fig:c10dense-tloss}
\vskip -0.1in
\caption{CIFAR10,DenseNetBC-40-12.}
\end{center}
\end{subfigure}%
\end{center}
\vskip -0.35in
\caption{ \label{fig:tloss}
  Additional examples of test loss curves of \oursL.  
  The arrows indicate the direction of time flow.
}
\vskip 0.4in
\centering
\begin{subfigure}[b]{0.2\linewidth}
\begin{center}
\centerline{\includegraphics[width=1\linewidth]{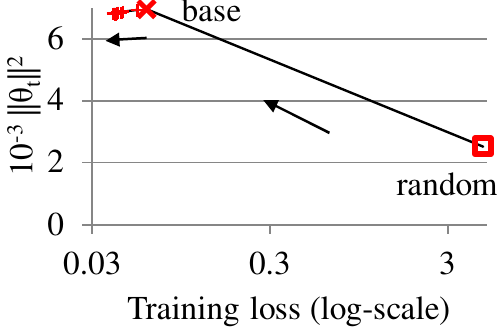}}
\vskip -0.1in
\caption{ \label{fig:c100k1-l2-a1}
\baseLoop\ ($\ssa$=1)
}
\end{center}
\end{subfigure}%
\begin{subfigure}[b]{0.2\linewidth}
\begin{center}
\centerline{\includegraphics[width=1\linewidth]{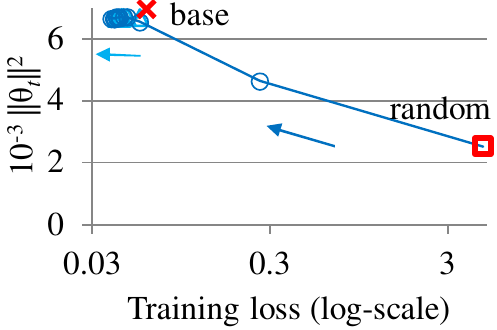}}
\vskip -0.1in
\caption{ \label{fig:c100k1-l2-a09}
  $\ssa$=0.9
}
\end{center}
\end{subfigure}%
\begin{subfigure}[b]{0.2\linewidth}
\begin{center}
\centerline{\includegraphics[width=1\linewidth]{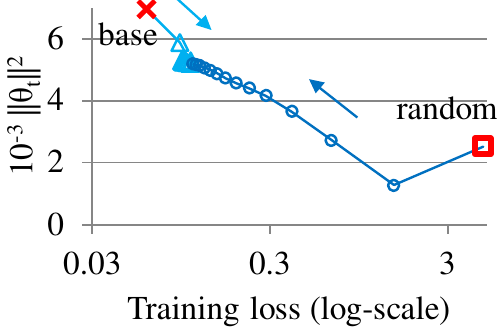}}
\vskip -0.1in
\caption{ \label{fig:c100k1-l2-a03}
  $\ssa$=0.3
}
\end{center}
\end{subfigure}%
\begin{subfigure}[b]{0.2\linewidth}
\begin{center}
\centerline{\includegraphics[width=1\linewidth]{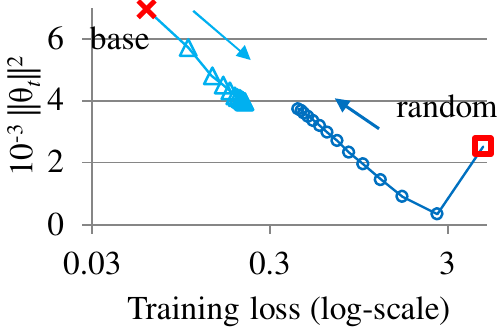}}
\vskip -0.1in
\caption{ \label{fig:c100k1-l2-a01}
  $\ssa$=0.1
}
\end{center}
\end{subfigure}%
\begin{subfigure}[b]{0.2\linewidth}
\begin{center}
\centerline{\includegraphics[width=1\linewidth]{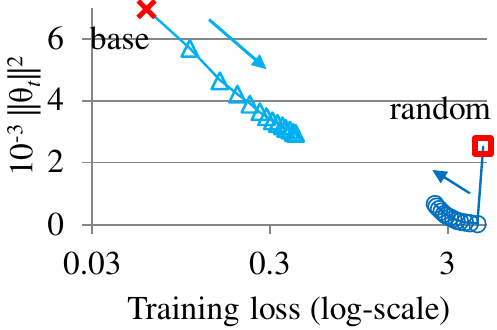}}
\vskip -0.1in
\caption{ \label{fig:c100k1-l2-a001}
  $\ssa$=0.01
}
\end{center}
\end{subfigure}%
\vskip -0.25in
\caption{ \label{fig:c100k1-l2-alpha}
Parameter size $\|\prm_t\|^2$ of \iniBase (`$\triangle$') and \iniRandom(`$\circ$').  
with five values of $\ssa$ (becoming smaller from left to right), 
in relation to training loss. \oursL.  $\tT$=25. 
CIFAR100.  ResNet-28. 
Matching figures with Figure \ref{fig:c100k1-tloss-alpha}.  
As $\ssa$ becomes smaller, 
the (potential) meeting point shifts 
further away from the base model. 
The left-most figure is \baseLoop, which is equivalent to 
$\ssa$=1. 
The arrows indicate the direction of time flow.  
}
\vskip -0.1in
\end{figure}

Figure \ref{fig:c100k1-joint-terr-with-legend} shows test error (\%) 
in relation to training loss with a small ResNet on CIFAR100. 
Additional examples of test-loss curves are shown in Figure \ref{fig:tloss}. 
Figure \ref{fig:c100k1-l2-alpha} shows the parameter size $\|\prm_t\|^2$ 
in relation to training loss, in the settings of Figure \ref{fig:c100k1-tloss-alpha} 
in the main paper.  

\begin{table}[h]
\begin{center} \begin{small} \begin{tabular}{|c|c|c|c|c|c|c|}
\multicolumn{2}{r}{Case\#} &\multicolumn{1}{c}{1}&\multicolumn{1}{c}{2}&\multicolumn{1}{c}{3}&\multicolumn{1}{c}{4}&\multicolumn{1}{c}{5}
\\
\hline
\multicolumn{2}{|c|}{Data} &\multicolumn{2}{|c|}{ large-Yelp} & \multicolumn{3}{|c|}{small-Yelp} \\
\hline
\multicolumn{2}{|c|}{Embedding learning?} & \multicolumn{1}{|c|}{Yes} & \multicolumn{1}{|c|}{No} 
                                          & \multicolumn{1}{|c|}{Yes} & \multicolumn{2}{|c|}{No} \\
\hline
\multicolumn{2}{|c|}{Loss function} & \multicolumn{4}{|c|}{cross-entropy} & $\dagger$  \\
\hline
\multirow{3}{*}{baselines} 
&\baseInTable   &     2.81 &     2.98 &     3.80&     5.43 &   5.32 \\ 
& \baseLoop     &     2.63 &     2.88 &     3.90&     5.43 &   5.34 \\
& w/ dropout    &     2.70 &     2.95 &     3.90&     5.34 &   5.35 \\
\hline
\multirow{3}{*}{\ours}   
& \iniRandom  &{\bf 2.34}&     2.72 &{\bf 3.70}&     5.06 &     5.00 \\
& \iniBase    &     2.38 &{\bf 2.70}&     3.77 &     5.15 &     4.98 \\
& \iniBase/2  &     2.43 &     2.74 &     3.73 &{\bf 4.99}&{\bf 4.96}\\
\hline
\end{tabular} \end{small} \end{center}
\vskip -0.1in
\caption{\label{tab:text-results}
  Test error (\%) on sentiment classification.  Median of 3 runs.
  7-block 250-dim DPCNN (10M parameters).  
  $\dagger$ Squared hinge loss.  
}
\vskip -0.1in 
\end{table}

\subsection{Additional experiments on text data}
We tested \ours\ on sentiment classification
to predict whether reviews are positive or negative, 
using the polarized Yelp dataset (\#train: 560K, \#test: 38K) \cite{ZZL15}.  
The best-performing models on this task are transformers pre-trained with language modeling on 
large and general text data such as Bert \cite{bert19} and XLnet \cite{xlnet19}.  
However, these models are generally large and time-consuming to train
using a GPU (i.e., without TPUs used in the original work).  
Therefore, instead, we used the deep pyramid convolutional neural network (DPCNN) 
\cite{dpcnn17} as our base model. 
In these experiments, we used \oursL.  

Table \ref{tab:text-results} shows the test error results in five settings.  
The last three use relatively small training sets of 45K data points 
and validation sets of 5K data points, randomly chosen from the original training set, 
while the first two use the entire training set (560K data points) except for 
5K data points held out for validation (meta-parameter tuning).  
DPCNNs optionally take additional features produced by embeddings of text regions that are trained with unlabeled data, 
similar to language modeling.  
Cases 1 and 3 exploited this option, training embeddings using the entire training set as unlabeled data; 
\ref{apx:text-details} below provides the details.  
As in the image experiments, we used the cross entropy loss with softmax except for 
Case 5, where the quadratic hinge loss $\loss_y(f) = \max(0,1-yf)^2$ for $y \in \{-1,1\}$ was used.  
This serves as an example of extending self-distillation (formulated specifically with the cross-entropy loss)
to general loss functions. 

In all the five settings, \ours\ achieves better test errors than the baseline methods, 
which shows the effectiveness of our approach in these settings.  
On this task, dropout turned out to be not very effective, 
which is, however, a reminder that the effectiveness of regularization methods can be data-dependent in general. 

\begin{table}[h]
\vskip -0.1in
\begin{center} \begin{small} \begin{tabular}{|c|l|c|c|c|c|}
  \multicolumn{2}{r}{Case\#}
 & \multicolumn{1}{c}{1} & \multicolumn{1}{c}{2} & \multicolumn{1}{c}{} \\
\hline
\multicolumn{2}{|c|}{}
  &\multicolumn{2}{|c|}{LM-like prep?}                        & Runtime &Text for \\ %
\cline{3-4}
\multicolumn{2}{|c|}{}
  &\multicolumn{1}{|c|}{$~\;$Yes$~\;$}&\multicolumn{1}{|c|}{$~\;$No$~\;$} & (sec/K)   &prep (GB)\\ %
\hline
(J \& Z, 2017)&DPCNN       &{\em 2.64}&{\em 3.30}&$~$0.1 &0.4\\
\hline
\multirow{2}{*}{This work}
&Table \ref{tab:text-results} best& 2.34 & 2.70 &$~$0.1 &0.4 \\
&Ensemble                 &{\bf 2.18}&{\bf 2.46}&$~$0.9 &0.4 \\
\hline
\multirow{2}{*}{\cite{bert19}}
                &Bert base &     2.25 &    6.19 &$~$5.7        &13 \\
                &Bert large&{\em 1.89}&   --    & 17.9         &13  \\
\hline
\multirow{2}{*}{\cite{xlnet19}}
                &XLnet base&     1.92 &    4.51 & 17.2         &13  \\
                &XLnet large&{\em 1.55}&  --    & 40.5         &126 \\
\hline
\end{tabular} \end{small} \end{center}
\vskip -0.1in
\caption{\label{tab:text-results2}
  \ours\ ensemble results on Yelp in comparison with previous models.
  Test error (\%) with or without embedding learning (DPCNN) or 
  language modeling-based pre-training (Bert and XLnet), respectively, 
  corresponding to Cases 1 \& 2 of Table \ref{tab:text-results}.  
  Runtime: real time in seconds for labeling 1K instances using a single GPU with 11GB device memory, 
  measured in the setting of Case 1; the average of 3 runs.  
  The last column shows 
  amounts of text data in giga bytes used for pre-training or embedding learning in Case 1.  \\
  The test errors in {\em italics} were copied from the respective publications 
  except that the Bert-large test error is from \cite{XDHLL19}; 
  other test errors and runtime were obtained by our experiments.  
  Our ensemble test error results are in bold.  
}
\end{table}

It is known that performance can be improved by making an ensemble of models from different stages 
of self-distillation, e.g., \cite{Ban18}.  
In Table \ref{tab:text-results2}, we report ensemble performances of 
DPCNNs trained with \ours, in comparison with the previous best models.  
Test errors with and without embedding learning 
(or language modeling-based pre-training for Bert and XLnet) are shown, 
corresponding to Cases 1 and 2 in Table \ref{tab:text-results}.  
The ensemble results were obtained by adding after applying softmax 
the output values of 20 DPCNNs (or 10 in Case 2) of last 5 stages 
of \ours\ training with different training options; details are provided in \ref{apx:text-details}. 

With embedding learning, the ensemble of DPCNNs trained with \ours\ 
achieved test error 2.18\%, which slightly beats 2.25\% of pre-trained Bert-base, 
while testing (i.e., making predictions) of this ensemble is more than 6 times faster than Bert-base, 
as shown in the `Runtime' column.  
(Note, however, that runtime depends on implementation and hardware/software configurations.)
That is, using \ours, we were able to obtain a classifier that is as accurate as and much faster 
than a pre-trained transformer.  

\cite{xlnet19} and \cite{XDHLL19} report 
1.55\% and 1.89\% 
using a pre-trained large transformer, XLnet-large and Bert-large, respectively.  
We observe that the runtime and the amounts of text used for pre-training 
(the last two columns) indicate that their high accuracies come with steep cost at every step: 
pre-training, fine-tuning, and testing.  
Compared with them, an ensemble of \ours-trained DPCNNs is a much lighter-weight solution 
with an appreciable accuracy.  
Also, our ensemble without embedding learning outperforms Bert-base and XLnet-base without pre-training, 
with relatively large differences (Case 2).  
A few attempts of training Bert-large and XLnet-large from scratch also resulted in 
underperforming DPCNNs, 
but we omit the results as we found it infeasible to complete meta-parameter tuning in reasonable time.  

On the other hand, 
it is plausible that the accuracy of the high-performance pre-trained transformers can be further improved 
by applying \ours\ to their fine-tuning, which would further push the state of the art.  
Though currently precluded by our computational constraints, 
this may be worth investigating in the future.  

\subsubsection{Details of the text experiments}
\label{apx:text-details}

\paragraph{Embedding learning}
It was shown in \cite{dpcnn17} that classification accuracy can be improved by 
training an embedding of small text regions (e.g., 3 consecutive words) 
for predicting neighboring text regions (`target regions') on unlabeled data 
(similar to language modeling) 
and then using the learned embedding function to produce additional features for the classifier.  
In this work, 
we trained the following two types of models with respect to use of embedding learning. 
\begin{itemize}
\item Type-0 did not use any additional features from embedding learning. 
\item Type-1 used additional features from the following two types of embedding simultaneously: 
  \begin{itemize}
  \item the embedding of 3-word regions as a function of a bag of words to a 250-dim vector, and 
  \item the embedding of 5-word regions as a function of a bag of word \{1,2,3\}-grams to a 250-dim vector. 
  \end{itemize}
\end{itemize}
Embedding training was done using the entire training set (560K reviews, 391MB) as unlabeled data
disregarding the labels.  

It is worth mentioning that our implementation of embedding learning differs from the 
original DPCNN work \cite{dpcnn17}, as a result of pursuing an efficient implementation in pyTorch 
(the original implementation was in C++). 
The original work used the bag-of-word representation for target regions 
(to be predicted) and minimized squared error with negative sampling. 
In this work we minimized the log loss without sampling where the target probability 
was set by equally distributing the probability mass among the words in the target regions. 

\paragraph{Table \ref{tab:text-results}}
Optimization was done by SGD.  
The learning rate scheduling of the base model and each stage of \baseLoop\ and 
\ours\ was fixed to 9 epochs with the initial learning rate $\eta$ followed by 
1 epoch with 0.1$\eta$. 
The mini-batch size was 32 for small training data and 128 
for large training data.  
We chose the weight decay parameter from \{1e-4, 2e-4, 5e-4, 1e-3\}
and the initial learning rate from \{0.25, 0.1, 0.05\}, 
using the validation data, 
except that for \ours\ on the large training data, 
we simply used the values chosen for the base model, 
which were weight decay 1e-4 and learning rate 0.1 (with embedding learning)
and 0.25 (without embedding learning).  

For \ours, we chose the number of stages $T$ from \{1,2,\ldots,25\} 
and $\ssa$ from \{0.3, 0.5\}, using the validation data.
$\ssa=0.5$ was chosen in most cases.  

\paragraph{Table \ref{tab:text-results2}}
The ensemble performances were obtained by combining 
\begin{itemize}
\item 
20 DPCNNs ( $T \in$ \{21, 22, \ldots, 25\} $\times$ \{\iniRandom, \iniBase\} $\times$ \{Type-0, Type-1\} )
  in Case 1, and 
\item 
10 DPCNNs ( $T \in$ \{21, 22, \ldots, 25\} $\times$ \{\iniRandom, \iniBase\} $\times$ \{Type-0\}) in Case 2.  
\end{itemize}
To make an ensemble, the model output values were added after softmax.  

\paragraph{Transformers}
The Bert and XLnet experiments were done using HuggingFace's Transformers\footnote{
  {\tt https://huggingface.co/transformers/} 
}
in pyTorch. 
Following the original work, 
optimization was done by Adam with linear decay of learning rate. 
For enabling and speeding up training using a GPU, 
we combined the techniques of gradient accumulation and variable-sized mini-batches (for improving parallelization) 
so that weights were updated after obtaining the gradients from approximately 128 data points. 
128 was chosen, following the original work. 
To measure runtime of transformer testing, we used variable-sized mini-batches for speed-up 
by improving the parallelism on a GPU.  

\end{document}